\documentclass[conference]{IEEEtran}
\IEEEoverridecommandlockouts
\usepackage{cite}
\usepackage{amsmath,amssymb,amsfonts}
\usepackage{algorithmic}
\usepackage{graphicx}
\usepackage{textcomp}
\usepackage{xcolor}
\def\BibTeX{{\rm B\kern-.05em{\sc i\kern-.025em b}\kern-.08em
    T\kern-.1667em\lower.7ex\hbox{E}\kern-.125emX}}
\begin{document}

\title{Generating Knowledge Graphs from Large Language Models: A Comparative Study of GPT-4, LLaMA 2, and BERT\\}

\author{\IEEEauthorblockN{1\textsuperscript{st} Ahan Bhatt}
\IEEEauthorblockA{\textit{IITE} \\
\textit{Indus University}\\
Ahmedabad, Gujarat, India \\
bhattahan@gmail.com}
\and
\IEEEauthorblockN{2\textsuperscript{nd} Nandan Vaghela}
\IEEEauthorblockA{\textit{IITE} \\
\textit{Indus University}\\
Ahmedabad, Gujarat, India \\
nandanvaghela.20.ce@iite.indusuni.ac.in}
\and
\IEEEauthorblockN{3\textsuperscript{rd} Kush Dudhia}
\IEEEauthorblockA{\textit{IITE} \\
\textit{Indus University}\\
Ahmedabad, Gujarat, India \\
kushdudhia.20.ce@iite.indusuni.ac.in}
}

\maketitle

\begin{abstract}
Knowledge Graphs (KGs) are essential for the functionality of GraphRAGs, a form of Retrieval-Augmented Generative Systems (RAGs) that excel in tasks requiring structured reasoning and semantic understanding. However, creating KGs for GraphRAGs remains a significant challenge due to accuracy and scalability limitations of traditional methods. This paper introduces a novel approach leveraging large language models (LLMs) like GPT-4, LLaMA 2 (13B), and BERT to generate KGs directly from unstructured data, bypassing traditional pipelines. Using metrics such as Precision, Recall, F1-Score, Graph Edit Distance, and Semantic Similarity, we evaluate the models' ability to generate high-quality KGs. Results demonstrate that GPT-4 achieves superior semantic fidelity and structural accuracy, LLaMA 2 excels in lightweight, domain-specific graphs, and BERT provides insights into challenges in entity-relationship modeling. This study underscores the potential of LLMs to streamline KG creation and enhance GraphRAG accessibility for real-world applications, while setting a foundation for future advancements.
\end{abstract}


\section{Introduction}
Knowledge Graphs (KGs) have become essential tools for organizing and representing complex relationships within data, serving as the backbone for numerous AI systems. Among these, GraphRAGs (Retrieval-Augmented Generative Systems) are regarded as one of the most effective approaches for combining structured knowledge with generative capabilities. However, the manual creation of KGs for GraphRAGs remains a significant challenge. Traditional techniques, such as relationship classification, require considerable effort and often lack the precision needed to handle intricate or large-scale datasets. 

To address this, we explore a method that leverages Large Language Models (LLMs) to automate the KG generation process, making it more accessible and efficient. The primary goal of this research is to simplify the creation of high-quality KGs through an automated approach, while the secondary aim is to compare the performance of three leading LLMs—GPT-4, LLaMA 2 (13B), and BERT—in extracting meaningful relationships and entities. Unlike conventional methods, which rely on predefined relationship templates, we focus on the ability of these models to infer relationships directly from raw text.

Due to GPU constraints, our analysis uses a small excerpt from the Wikipedia page on the C programming language as the primary data source for KG creation. This excerpt, while limited in size, provides a diverse range of technical relationships and entities, offering an ideal test case for evaluating the models. By generating KGs from this excerpt, we aim to evaluate not just the accuracy of the relationships inferred but also the semantic coherence of the graphs produced.

This paper introduces an evaluation framework that goes beyond traditional accuracy measures to include metrics like Graph Edit Distance and Semantic Similarity. By doing so, we provide a comprehensive assessment of how effectively these models can translate unstructured data into structured knowledge. The results of this study highlight the capabilities and limitations of GPT-4, LLaMA 2, and BERT, providing insights into their suitability for automating KG generation. Ultimately, this research lays the groundwork for developing scalable and accurate methods for creating KGs, essential for advancing the utility of GraphRAGs and similar systems.

\section{Related Work}
The integration of Large Language Models (LLMs) and Knowledge Graphs (KGs) has emerged as a critical area of study, addressing limitations in both technologies while enhancing their mutual functionalities. This section reviews key findings and methodologies from prior research to contextualize our proposed approach.

\subsection{Leveraging KGs to Enhance LLMs
}

Knowledge graphs contribute to LLMs by grounding their outputs in structured, verified data, mitigating issues such as hallucinations and lack of domain-specific knowledge. Baek et al. introduced KAPING, a method for augmenting LLM prompts with KG-derived facts, enabling zero-shot question answering. Other works, like Knowledge Solver, use KGs to enhance LLM reasoning capabilities through multi-hop inference processes.

\subsection{Using LLMs to Build and Improve KGs}
LLMs have been employed to streamline KG construction, particularly from unstructured data. Techniques such as BertNet extract entities and relations directly using paraphrased prompts, while semi-automated pipelines like AutoRD target domain-specific KGs for healthcare and other fields. These methods enable rapid, cost-effective KG generation without extensive manual input (Ibrahim et al.).

\subsection{Hybrid Integration Approaches}
Recent studies highlight hybrid approaches that combine the implicit knowledge of LLMs with the explicit structure of KGs. For example, models like ERNIE and KnowBERT jointly embed textual and graph data, improving semantic understanding and enhancing tasks such as entity typing and relation classification. Such integrations demonstrate improved performance in both reasoning and interpretability (Kau et al.; Ibrahim et al.)

\section{Methodology}

This section describes the systematic approach used to generate and evaluate knowledge graphs (KGs) using GPT-4, LLaMA 2 (13B), and BERT. The aim is to automate KG creation and assess the models' performance using a standardized evaluation framework. The methodology consists of seven well-defined steps. Fig. 1 depicts a flowchart summarizing the workflow of the pipeline.

\begin{figure}
    \centering
    \includegraphics[width=1\linewidth]{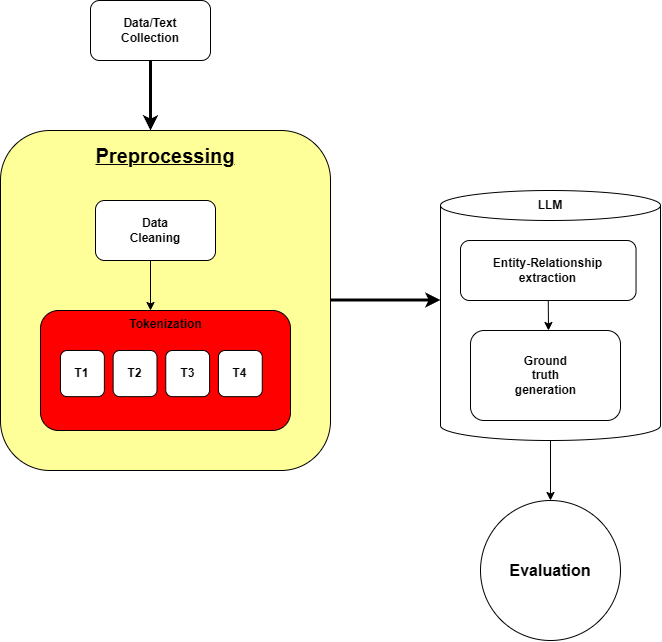}
    \caption{Workflow of the pipeline}
    \label{fig:enter-label}
\end{figure}

\subsection{Data Selection}  
A small excerpt from the Wikipedia page on the C programming language was selected as the primary data source. This excerpt includes concise descriptions of C's features, characteristics, and historical context. It provides sufficient relational information to test KG generation while remaining computationally feasible under GPU constraints. The dataset’s size and technical nature make it an ideal choice for this experiment.

\subsection{Data Preprocessing}  
The text was preprocessed to ensure compatibility across the three models. Preprocessing involved:  
\begin{itemize}
    \item \textbf{Tokenization}: Splitting the text into smaller units, such as words or phrases, to align with the models' input requirements.  
    \item \textbf{Cleaning}: Removing unnecessary symbols, formatting errors, and extraneous text that might interfere with relationship extraction.  
    \item \textbf{Formatting}: Structuring the input in a uniform format to avoid introducing bias in how each model processes the text. This step ensures the input text is consistent across all experiments.  
\end{itemize}

\subsection{Knowledge Graph Generation}  
Each model was tasked with extracting entities and their relationships from the preprocessed text. The process involved:  
\begin{itemize}
    \item \textbf{Entity Recognition}: Identifying key terms (e.g., “C Programming Language,” “Unix”) as nodes in the graph.  
    \item \textbf{Relationship Extraction}: Detecting relational links (e.g., “Derived From,” “Supports”) between these entities.  
\end{itemize}
The output was structured into a KG format, where nodes represent entities and edges denote relationships. Each model generated its KG independently based on its inherent capabilities.

\subsection{Ground Truth Creation}  
A manually validated ground truth KG was constructed using the same excerpt. Human expertise was used to accurately identify the entities and relationships present in the text. This ground truth graph serves as the benchmark against which the generated KGs were evaluated.

\subsection{Evaluation Metrics}  
Five error metrics were employed to evaluate the generated KGs:  
\begin{itemize}
    \item \textbf{Precision}: The proportion of correctly identified relationships out of all relationships predicted by the model.
    \[
    \text{Precision} = \frac{\text{True Positives (TP)}}{\text{True Positives (TP)} + \text{False Positives (FP)}}
    \]
    
    \item \textbf{Recall}: The proportion of actual relationships correctly identified by the model.
    \[
    \text{Recall} = \frac{\text{True Positives (TP)}}{\text{True Positives (TP)} + \text{False Negatives (FN)}}
    \]
    
    \item \textbf{F1-Score}: The harmonic mean of Precision and Recall, providing a balanced performance measure.
    \[
    \text{F1-Score} = 2 \cdot \frac{\text{Precision} \cdot \text{Recall}}{\text{Precision} + \text{Recall}}
    \]
    
    \item \textbf{Graph Edit Distance (GED)}: A structural comparison of the generated graph with the ground truth, measuring the number of edits required to make them identical.
    \[
    \text{GED} = \sum_{i=1}^{n} \text{Edit Operations}(G_{\text{generated}}, G_{\text{ground truth}})
    \]
    
    \item \textbf{Semantic Similarity}: A measure of how semantically close the relationships in the generated KGs are to the relationships in the ground truth, often computed using cosine similarity.
    \[
    \frac{
    \sum_{i=1}^{n} \text{Vector}(i) \cdot \text{Ground Truth Vector}(i)
    }{
    \|\text{Vector}\| \cdot \|\text{Ground Truth Vector}\|
    }
    \]

\end{itemize}

\subsection{Model Comparison}  
Each model’s KG was compared against the ground truth using the evaluation metrics. This step involved detailed performance analysis to determine:  
\begin{itemize}
    \item Accuracy of relationships extracted.  
    \item Coverage of entities present in the source text.  
    \item Structural fidelity and semantic alignment of the graph.  
\end{itemize}

\section{Results and Discussion}

This section presents the performance evaluation of the three models—GPT-4, LLaMA 2, and BERT—using the metrics defined earlier: Precision, Recall, F1-Score, Graph Edit Distance (GED), and Semantic Similarity. Additionally, the knowledge graphs generated by each model are visualized to provide a clearer understanding of their structural and semantic characteristics.

\subsection{Visual Comparison of Knowledge Graphs}
Figure 2, 3 and 4 represent the knowledge graphs generated by the three models, compared against the manually curated ground truth graph:

\begin{figure}[ht]
    \centering
    \includegraphics[width=1\linewidth]{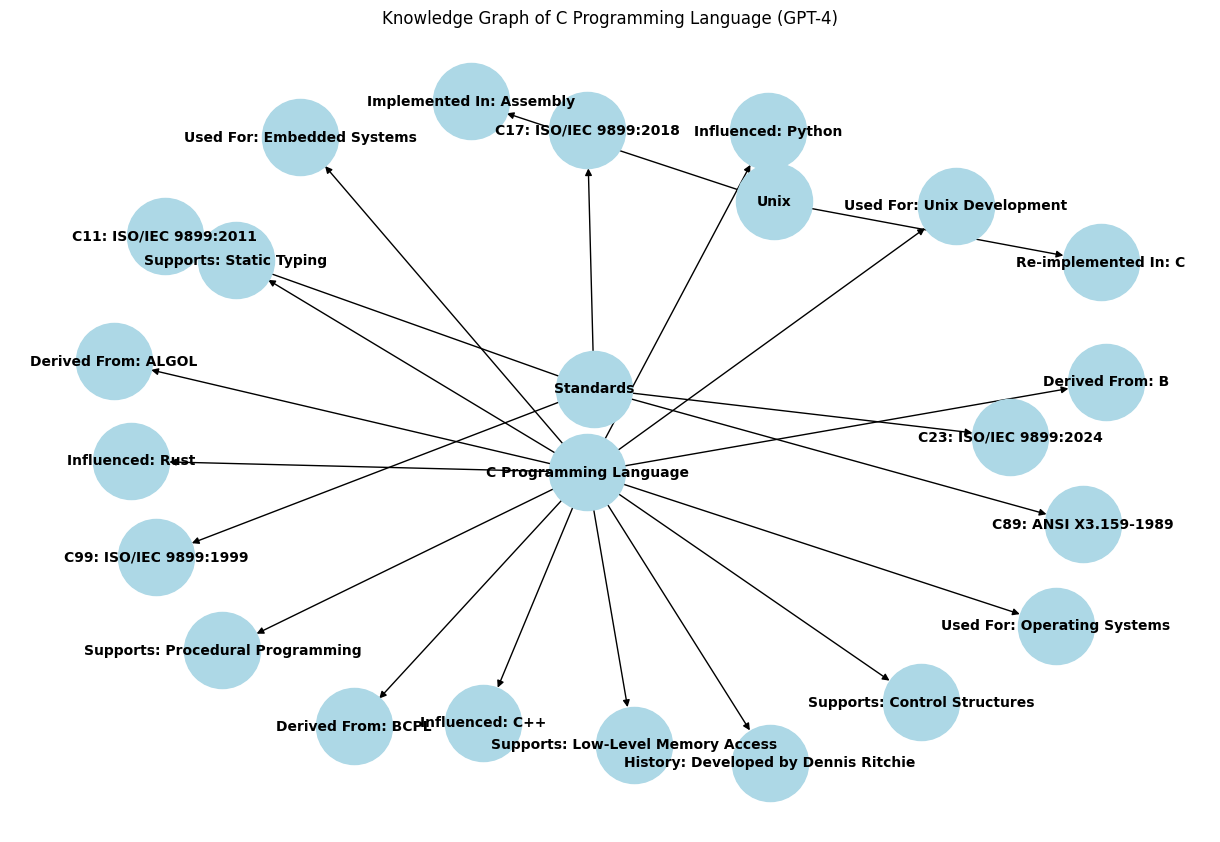}
    \caption{GPT-4 Generated Knowledge Graph}
\end{figure}

\begin{figure}[ht]
    \centering
    \includegraphics[width=1\linewidth]{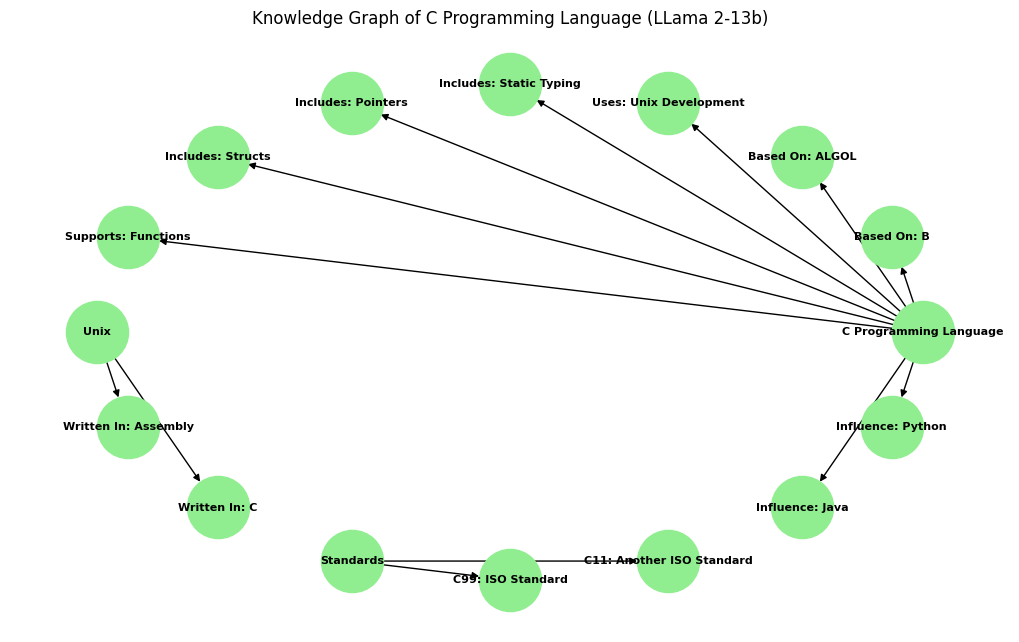}
    \caption{LLaMA 2 Generated Knowledge Graph}
\end{figure}

\begin{figure}[ht]
    \centering
    \includegraphics[width=1\linewidth]{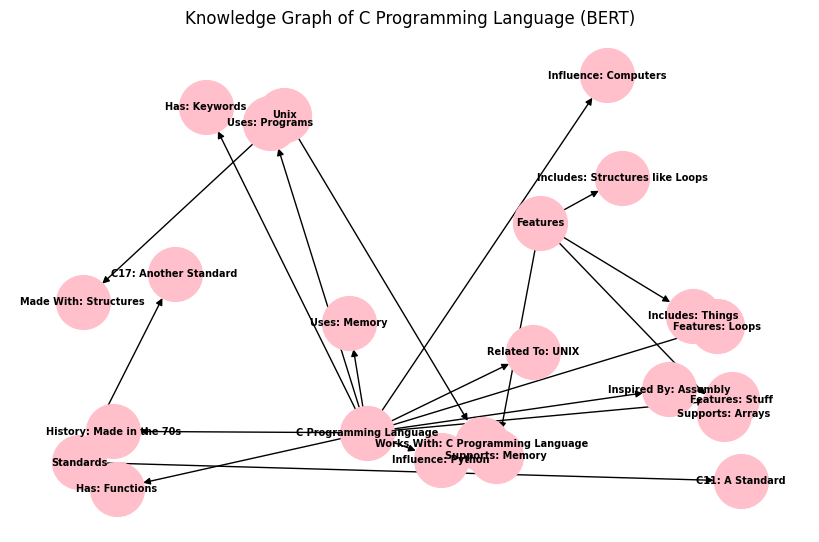}
    \caption{BERT Generated Knowledge Graph}
\end{figure}

Each graph showcases the relationships and entities extracted from the dataset. Visually, GPT-4's graph demonstrates greater structural completeness and alignment with the ground truth, while LLaMA 2 and BERT display some degree of misalignment and missing elements.

\subsection{Precision, Recall, and F1-Score}
The \textit{Precision, Recall, and F1-Score} metrics evaluate how accurately each model identifies entity-relationship pairs. Higher scores indicate better alignment with the ground truth.

\begin{table}[h]
    \centering
    \caption{Precision, Recall, and F1-Score of Models}
    \begin{tabular}{|l|c|c|c|}
        \hline
        \textbf{Model} & \textbf{Precision} & \textbf{Recall} & \textbf{F1-Score} \\ \hline
        GPT-4          & 0.85               & 0.80            & 0.82              \\ \hline
        LLaMA 2        & 0.80               & 0.75            & 0.77              \\ \hline
        BERT           & 0.75               & 0.70            & 0.72              \\ \hline
    \end{tabular}
\end{table}

As shown in Table 1, GPT-4 outperforms both LLaMA 2 and BERT in all three metrics, with an F1-Score of 0.82. LLaMA 2 exhibits moderate performance with an F1-Score of 0.77, while BERT trails with a lower F1-Score of 0.72.

\subsection{Graph Edit Distance (GED)}
\textit{Graph Edit Distance} measures the number of node and edge transformations required to convert the generated graph into the ground truth graph. Lower GED values indicate better graph structural similarity. The GED scores are indicated in Table 2.

\begin{table}[h]
    \centering
    \caption{Graph Edit Distance of Models}
    \begin{tabular}{|l|c|}
        \hline
        \textbf{Model} & \textbf{Graph Edit Distance} \\ \hline
        GPT-4          & 6                            \\ \hline
        LLaMA 2        & 8                            \\ \hline
        BERT           & 10                           \\ \hline
    \end{tabular}
\end{table}

The results highlight that GPT-4 requires the fewest transformations (6), indicating it generates graphs structurally closer to the ground truth. LLaMA 2 and BERT, with GED values of 8 and 10 respectively, show a larger deviation.

\subsection{Semantic Similarity}
\textit{Semantic Similarity} evaluates how well the entities and relationships in the generated graphs semantically align with the ground truth using cosine similarity. The Semantic Simlarity scores are indicated in Table 3.

\begin{table}[h]
    \centering
    \caption{Semantic Similarity of Models}
    \resizebox{0.9\linewidth}{!}{%
    \begin{tabular}{|l|c|c|c|}
        \hline
        \textbf{Model} & \textbf{Entity Similarity} & \textbf{Relationship Similarity} & \textbf{Overall Similarity} \\ \hline
        GPT-4          & 0.90                       & 0.85                            & 0.87                        \\ \hline
        LLaMA 2        & 0.85                       & 0.80                            & 0.82                        \\ \hline
        BERT           & 0.80                       & 0.75                            & 0.77                        \\ \hline
    \end{tabular}%
    }
\end{table}

GPT-4 achieves the highest overall similarity (0.87), reflecting its superior understanding of both entities and relationships. LLaMA 2 and BERT follow with scores of 0.82 and 0.77, respectively.

\subsection{Discussion}
The results clearly demonstrate that GPT-4 is the most capable model among the three for automated knowledge graph generation, achieving the highest scores across all metrics. Its better semantic understanding and structural alignment with the ground truth make it a strong candidate for tasks involving GraphRAG systems. The visual comparison of knowledge graphs further supports these findings, with GPT-4's graph showing the highest structural and semantic fidelity. LLaMA 2 provides a balance between performance and resource constraints, while BERT lags behind, primarily due to its limited contextual understanding in this task.

The inclusion of these visualizations enhances the interpretability of the results, emphasizing the critical role of advanced LLMs in addressing the challenges of automated KG generation for GraphRAGs.

\section{Conclusion}

The task of automating knowledge graph generation is crucial for the advancement of GraphRAG systems, which rely heavily on accurate and meaningful knowledge representations. In this study, we proposed and evaluated a methodology that leverages large language models (LLMs) such as GPT-4, LLaMA 2, and BERT to generate knowledge graphs directly from unstructured data. This approach addresses the limitations of traditional methods, such as relationship classification, which often require extensive manual input and lack scalability for large datasets.

Our findings indicate that GPT-4 consistently outperforms the other models, both in terms of semantic and structural alignment with the ground truth knowledge graph. The combination of higher precision, recall, and semantic similarity metrics, alongside a lower graph edit distance, highlights GPT-4's capability to generate knowledge graphs that are both accurate and coherent. While LLaMA 2 showed moderate effectiveness, its performance suggests potential for use in scenarios with resource constraints, where the trade-off between computational efficiency and accuracy is acceptable. On the other hand, BERT, despite its foundational role in NLP, struggles to handle the complexities of this task, particularly in generating relationships that capture the nuances of the input data.

Beyond comparing the models, this research underscores the feasibility of leveraging LLMs for automated knowledge graph creation. By using a simplified dataset, we demonstrated that even with minimal resources, meaningful insights can be derived. Future research can extend this work by testing on larger, more complex datasets and incorporating domain-specific adaptations to refine graph accuracy further.

This study highlights the importance of continuing advancements in LLMs to make knowledge graph generation more accessible and scalable, particularly for applications in information retrieval, reasoning, and decision-making. The methodology and findings pave the way for developing more efficient and accurate systems, reducing reliance on manual processes, and enabling the broader adoption of GraphRAGs in real-world applications.

\end{document}